\documentclass[10pt,twocolumn,letterpaper]{article}

\usepackage[pagenumbers]{cvpr} % To force page numbers, e.g. for an arXiv version

\usepackage{amsmath}
\newcommand\numberthis{\addtocounter{equation}{1}\tag{\theequation}}

\usepackage{chngcntr}
\usepackage[utf8]{inputenc}

\usepackage{enumitem}

\usepackage{wrapfig}
\usepackage{fontawesome5}
\usepackage{cuted}

\usepackage{xcolor}
\definecolor{mutedblue}{HTML}{0077BB} % color-blind-safe
\definecolor{mutedorange}{HTML}{EE7733} % color-blind-safe

\usepackage{pifont}
\usepackage{makecell}
\usepackage{booktabs}
\usepackage{multirow}

\usepackage{tikz}
\usepackage{changepage}
\usepackage{tikz-3dplot}
\usepackage{tikzscale}
\usetikzlibrary{plotmarks}
\usepackage{pgfplots}
\usetikzlibrary{fadings}
\usetikzlibrary{tikzmark,calc} 
\usetikzlibrary{shapes.arrows}
\usetikzlibrary{decorations.pathreplacing, decorations.markings}

\pgfplotsset{compat=1.18}

\makeatletter
\def\NAT@def@citea{\def\@citea{\NAT@separator}}
\makeatother

\definecolor{cvprblue}{rgb}{0.21,0.49,0.74} 
\usepackage[pagebackref,breaklinks,colorlinks,allcolors=cvprblue,hypertexnames=false]{hyperref}

\definecolor{colortrain}{rgb}{0.694, 0.796, 0.659} % (177,203,168)/255
\definecolor{colora}{rgb}{0.373,0.573,0.537} % (95,146,137)/255
\definecolor{colorb}{rgb}{0.294,0.463,0.627} % (75,118,160)/255
\definecolor{colorc}{rgb}{0.725,0.667,0.549} % (185,170,141)/255
\definecolor{colord}{rgb}{0.282,0.337,0.549} % (72,86,140)/255
\def\paperID{} % *** Enter the Paper ID here
\def\confName{CVPR}
\def\confYear{2026}

\title{USB: Unified Synthetic Brain Framework for\\Bidirectional Pathology–Healthy Generation and Editing}

\author{Jun Wang \qquad Peirong Liu \vspace{0.15cm} \\
Department of Electrical and Computer Engineering,\\
Data Science and AI Institute,\\
Johns Hopkins University \vspace{0.15cm}
\\ \tt\small \{jwang674, pliu53\}@jh.edu
}

\begin{document}

\maketitle

\begin{strip}
\centering
\vspace{-1.4cm}
\includegraphics[width=1.0\textwidth]{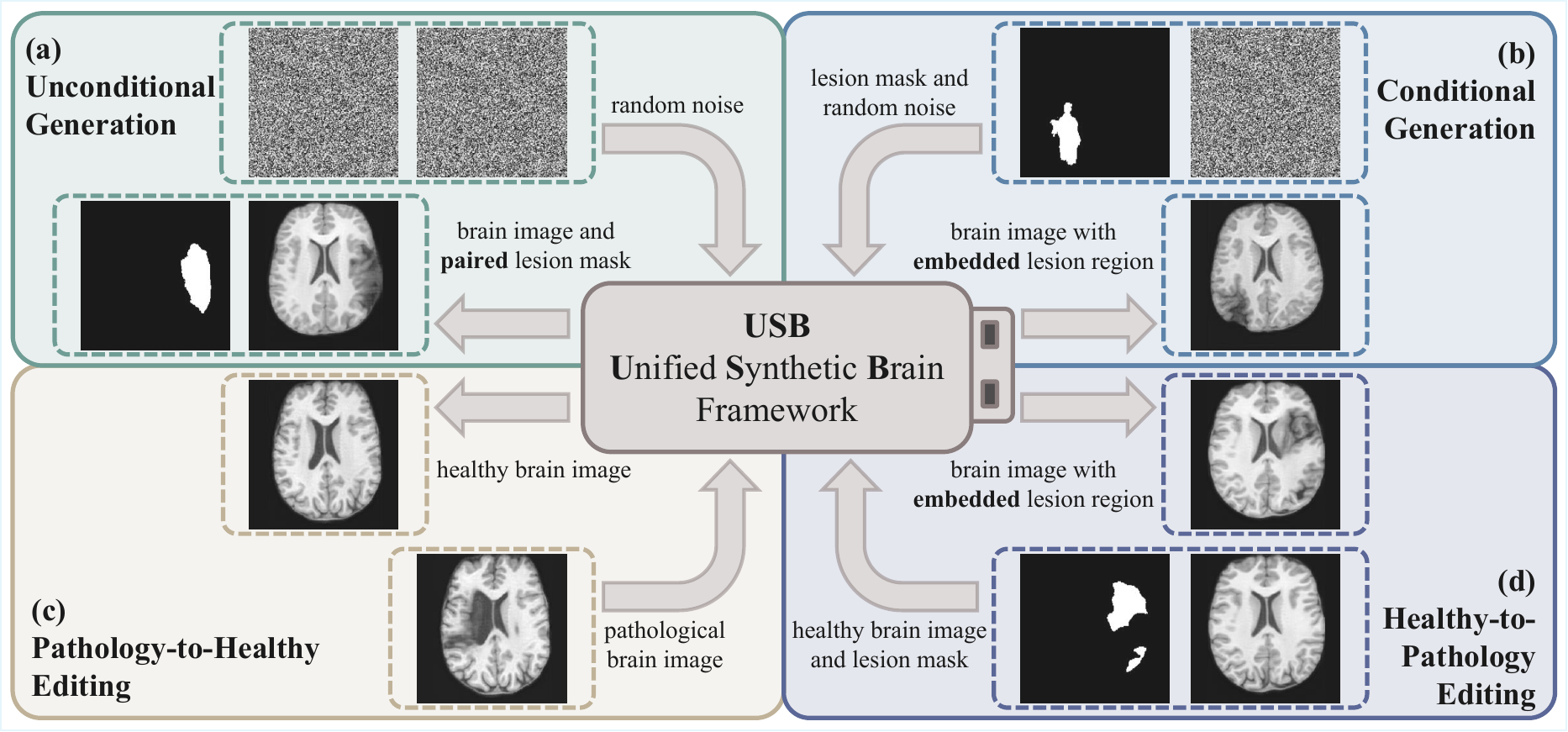}
\vspace{-0.6cm}
\captionof{figure}{\textbf{Four unified tasks supported by \texttt{USB}.} 
\textbf{\textcolor{colora}{(a)~Unconditional generation}} of paired brain image and lesion mask;  
\textbf{\textcolor{colorb}{(b)~Conditional generation}} given a specific lesion mask; 
\textbf{\textcolor{colorc}{(c)~Pathology-to-healthy editing}} that reconstructs a lesion-containing brain into a healthy brain;  
\textbf{\textcolor{colord}{(d)~Healthy-to-pathology editing}} that embeds a given lesion mask into a healthy brain.}
\label{fig:task}
\vspace{-0.2cm}
\end{strip}

\begin{abstract}
Understanding the relationship between pathological and healthy brain structures is fundamental to neuroimaging, connecting disease diagnosis and detection with modeling, prediction, and treatment planning. 
However, paired pathological–healthy data are extremely difficult to obtain, as they rely on pre- and post-treatment imaging, constrained by clinical outcomes and longitudinal data availability.
Consequently, most existing brain image generation and editing methods focus on visual quality yet remain domain-specific, treating pathological and healthy image modeling independently. 
We introduce \texttt{\textbf{USB}} (\texttt{\textbf{U}}nified \texttt{\textbf{S}}ynthetic \texttt{\textbf{B}}rain), the first end-to-end framework that unifies bidirectional generation and editing of pathological and healthy brain images. \texttt{USB} models the joint distribution of lesions and brain anatomy through a paired diffusion mechanism and achieves both pathological and healthy image generation. A consistency guidance algorithm further preserves anatomical consistency and lesion correspondence during bidirectional pathology-healthy editing. Extensive experiments on six public brain MRI datasets including healthy controls, stroke, and Alzheimer's patients, demonstrate \texttt{USB}’s ability to produce diverse and realistic results. 
By establishing the first unified benchmark for brain image generation and editing, \texttt{USB} opens opportunities for scalable dataset creation and robust neuroimaging analysis.\footnote{Code is available at \url{https://github.com/jhuldr/USB}.}

\end{abstract}

\section{Introduction}
\label{sec:intro}

Magnetic resonance imaging (MRI) provides noninvasive visualization of the human brain with exquisite and tunable soft-tissue contrast, serving as a cornerstone of neuroimaging and a primary tool for diagnosing and monitoring a wide range of neurological and cerebrovascular diseases~\cite{BrantZawadzki1992MPRA}. 
When interpreting brain MRIs, clinicians intuitively engage in \textit{bidirectional} and \textit{counterfactual reasoning}: from a pathological scan, they infer the likely healthy appearance if the patient were successfully treated; from a healthy scan, they project potential pathological outcomes given lesion patterns..
Such reasoning, developed through extensive clinical experience, mirrors the bidirectional generation and editing between pathological and healthy domains, 
and can be organized into two functional categories: \textit{(i)~generating} pathological and healthy brain images, and \textit{(ii)~editing} existing scans to embed or remove lesions.

Developing a model that can realize these capabilities, however, is far from trivial. 
First, acquiring \textit{paired} pathological–healthy data is highly challenging, as it requires pre- and post-treatment imaging that depends on successful clinical outcomes and longitudinal scan availability~\cite{bi2019artificial,cho2024prediction}. 
Second, although scans are widely available clinically, robust pathological analysis still depends heavily on large collections of high-quality, expert-annotated data~\cite{chen2025scaling,hasei2024high}, which are labor-intensive, non-reproducible, and disease-specific, while data privacy regulations further restrict data sharing. 
These challenges inherently hinder scalable modeling of the bidirectional connection between pathological and healthy brain anatomy, especially given the delicate and highly complex structure of the human brain.

Despite recent progress in brain image synthesis and pathology editing, existing methods remain task-specific and cannot bridge the bidirectional relationship between pathological and healthy domains. 
The generation schemes in SynthSR~\cite{iglesias2023synthsr} and Brain-ID~\cite{liu2024brainid} essentially perform \textit{healthy-to-healthy} synthesis under normal anatomical assumptions, fundamentally constraining their generalization to pathological scans.
PEPSI~\cite{liu2024pepsi} focuses exclusively on \textit{pathology-to-pathology} cross-modality synthesis and requires paired lesion annotations as input. More recent frameworks, UNA~\cite{liu2025una} and CondDiff~\cite{lawry2025conditional}, introduce \textit{single-direction} reconstruction of healthy images from pathological ones while relaxing explicit anatomy-lesion pairing. However, their oversimplified linear lesion encoding often leads to unrealistic pathological patterns~(Fig.~\ref{fig:editing_compare}) in the synthesized training samples, inherently limiting the model’s effectiveness for downstream applications, especially on real-world clinical data. 
These limitations underscore the need for a model that transitions between pathological and healthy domains, to enable seamless generation and editing without reliance on paired data or annotations.

Our goal is to develop a unified framework that formalizes clinicians’ bidirectional reasoning for data generation and pathology understanding, through an integrated generative model that generates and translates brain images across pathological and healthy domains. To this end, we introduce \texttt{\textbf{USB}}: the \textbf{\texttt{U}}nified \textbf{\texttt{S}}ynthetic \textbf{\texttt{B}}rain framework for
bidirectional pathology–healthy generation and editing of brain images. 
Our main contributions are summarized as follows:

\begin{itemize}
\item  
As shown in Fig.~\ref{fig:task}, \texttt{USB} is the \textit{first} framework to achieve \textbf{\textcolor{colora}{(a)}}~unconditional paired lesion–brain image generation, \textbf{\textcolor{colorb}{(b)}}~conditional brain image generation from lesion masks, \textbf{\textcolor{colorc}{(c)}}-\textbf{\textcolor{colord}{(d)}}~bidirectional editing between pathological and healthy brain images --- all within a single end-to-end architecture, \textit{without} the need for task-specific retraining.

\item 
We introduce a paired diffusion mechanism for bidirectional generation that jointly models lesion regions and brain anatomy, enabling anatomically consistent, paired lesion–brain generation.

\item 
We further propose a consistency guidance algorithm for bidirectional pathology–healthy editing, which preserves the original anatomy while ensuring accurate lesion–mask correspondence.

\item 
We conduct extensive experiments on \textit{six} public brain MRI datasets, spanning healthy subjects as well as stroke and Alzheimer’s disease cohorts. \texttt{USB} consistently achieves state-of-the-art performance across individual tasks and, more importantly, establishes the first unified benchmark for bidirectional pathology–healthy brain image generation-editing, enabling scalable dataset creation and robust neuroimaging analysis.

\end{itemize}

\section{Related Works}
\label{sec:related}

\subsection{Generative Models for Medical Imaging}

Generative modeling for medical imaging has rapidly advanced in recent years, motivated by the scarcity of annotated medical image data. 
Early efforts~\cite{singh2021medical,sun2022hierarchical,hong20213d,kebaili2023end,zong2018deep,baur2021autoencoders} based on GANs~\cite{goodfellow2014generative} and VAEs~\cite{kingma2013auto}  
demonstrated the feasibility of volumetric MRI/CT synthesis, yet suffered from instability, limited diversity, and blurred reconstructions at high resolution.
Recently, diffusion-based generative models, such as Denoising Diffusion Probabilistic Models (DDPMs)~\cite{ho2020denoising} and Latent Diffusion Models (LDMs)~\cite{rombach2022ldm}, have emerged as powerful and stable alternatives, achieving superior generative quality across diverse medical imaging tasks~\cite{kazerouni2023diffusion,xie2022measurement,kim2022diffusem, yang2023diffmic,fernandez2022can,zhang2024phy,peng2022towards, trippe2023diffusion}. 
For instance, DiffuseMorph~\cite{kim2022diffusem} integrates a deformation module into a diffusion network for brain and cardiac MRI registration; DiffMIC~\cite{yang2023diffmic} captures global–local priors for medical image classification; BrainSPADE~\cite{fernandez2022can} synthesizes brain label maps via a VAE–diffusion hybrid; and Phy-Diff~\cite{zhang2024phy} introduces physics-guided constraints for MRI generation.

Despite these advances, most existing diffusion-based methods remain \textit{task-specific} and thus difficult to generalize, or rely on pretrained models that \textit{require additional fine-tuning} for different downstream objectives. We instead aim to develop a unified generative framework capable of simultaneously handling diverse generation needs, including healthy scans, pathological scans, as well as corresponding lesion annotations, thereby unlocking the full potential of diffusion-based generative modeling in medical imaging.

%-------------------------------------------------------------------------
\subsection{Data-Driven Medical Image Editing}

Data-driven medical image editing aims to modify existing scans to improve clinical readability or support research applications, encompassing a broad range of tasks such as restoration~\cite{jiang2025restore,chung2022improving}, super-resolution~\cite{saharia2022image,iglesias2023synthsr,liu2024brainid}, and denoising~\cite{chung2022mr,xiang2023ddm,pfaff2024no}. 
SynthSR~\cite{iglesias2023synthsr} achieves contrast- and resolution-agnostic volumetric super-resolution. 
Brain-ID~\cite{liu2024brainid} learns contrast-agnostic feature representations that generalize effectively to super-resolution tasks. 
DDM$^2$~\cite{xiang2023ddm} introduces a self-supervised denoising strategy within a diffusion framework to enhance MRI quality. 
However, these editing-related models focus on the resulting visual quality itself, \textit{without} altering the underlying tissue's clinical meaning, (i.e., whether it is pathological or healthy). 

More recently, UNA~\cite{liu2025una} leverages synthesized pathological inputs for training, enabling pathology-to-healthy reconstruction. However, its oversimplified linear pathology encoding strategy often produces unrealistic lesion textures in its synthesized images~(Fig.~\ref{fig:editing_compare}~(b)), inherently limiting the model's capacity when applied to real-world clinical scans. 
LeFusion~\cite{zhang2024lefusion} proposes a diffusion-based learning objective emphasizing lesion regions, and achieves superior performance in synthesizing and detecting cardiac lesions and lung nodules. 
Nevertheless, this disease- and lesion-focused design prevents the model's generalization to healthy scans, and broader image analysis tasks beyond lesion segmentation. 
Most importantly, all these approaches remain confined to \textit{single-direction} transformations and can not perform bidirectional editing between pathological and healthy domains. 
In contrast, we aim for a unified, bidirectional framework that supports both pathology-to-healthy reconstruction and healthy-to-pathology synthesis for brain images, within a single end-to-end architecture.

\begin{figure*}[t]
\centering
\includegraphics[width=0.95\linewidth]{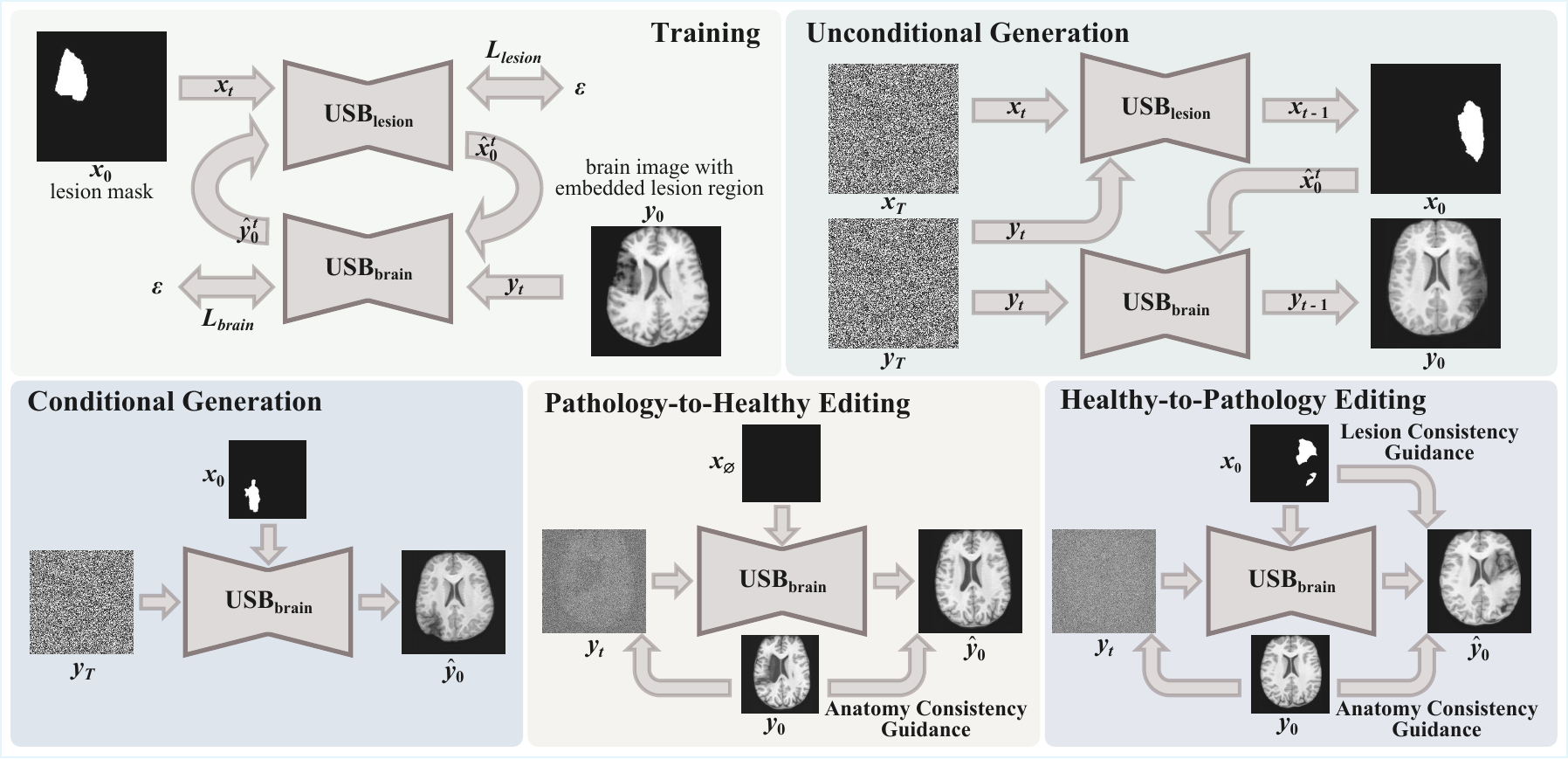}
\vspace{-0.2cm} 
\caption{\textbf{Framework of \texttt{USB}, illustrating the overall architecture of our unified synthetic brain framework.} 
\textcolor{colortrain}{(top-left)} The bidirectional training process of \texttt{USB}.  
\textcolor{colora}{(top-right)} Unconditional generation of paired lesion–brain image.  
\textcolor{colorb}{(bottom-left)} Conditional brain image generation given a lesion mask.  
\textcolor{colorc}{(bottom-middle)} Pathology-to-healthy editing given a pathology brain image with lesion embedded.  
\textcolor{colord}{(bottom-right)} Healthy-to-pathology editing given a healthy brain image and a lesion mask. 
}
\label{fig:framework}
\vspace{-0.3cm}
\end{figure*}

\section{Unified Synthetic Brain Framework}
\label{sec:method}

Fig.~\ref{fig:framework} provides an overview of \texttt{USB}'s framework, which integrates bidirectional training and cohesively supports four individual tasks: unconditional paired lesion–brain generation, conditional brain generation given lesion masks, and bidirectional pathology-healthy brain image editing.

In Sec.~\ref{subsec:generation}, we first introduce \texttt{USB}'s paired diffusion design, which enables the bidirectional generation of brain images and lesion masks within a unified diffusion framework. 
In Sec.~\ref{subsec:editing}, we further generalize \texttt{USB} to support bidirectional editing tasks (pathology-to-healthy, healthy-to-pathology) where we propose anatomy and lesion consistency guidance mechanisms that preserve structural fidelity and ensure lesion correspondence throughout editing.

\subsection{Paired Diffusion for Bidirectional Generation} 
\label{subsec:generation}

To model the joint distribution of a lesion mask $x_0$, and its corresponding brain image $y_0$ embedding $x_0$, we extend the standard diffusion framework~\cite{rombach2022ldm} to a paired diffusion formulation supporting bidirectional generation. As shown in Fig.~\ref{fig:framework}~\textcolor{colortrain}{(top-left)}, \texttt{USB} integrates the co-evolution of two sub-models that interact coherently during training: \textit{(i)}~\texttt{USB}\textsubscript{brain} predicts brain image conditioned on the lesion estimate $\hat x_0$ from \texttt{USB}\textsubscript{lesion}; and \textit{(ii)}~\texttt{USB}\textsubscript{lesion} predicts lesion masks conditioned on the image estimate $\hat y_0$ from \texttt{USB}\textsubscript{brain}. 

Let $\{x_t, y_t\}_{t=1}^T$ denote the noisy latent variables of lesion masks and brain image at diffusion timestep $t$. We reformulate the forward and reverse diffusion processes as follows:

\vspace{0.1cm}
\noindent \textbf{Paired Forward Diffusion.}
Given a lesion-brain pair $(x_0, y_0)$ sampled from the empirical distribution $q(x_0, y_0)$, the forward diffusion process gradually perturbs both inputs with independent Gaussian noise:
\vspace{-0.2cm}
\begin{align}
q(x_t \mid x_{t-1}) &= \mathcal{N}\big(x_t; \sqrt{1-\beta_t}\, x_{t-1}, \, \beta_t \mathbf{I}\big),\\
q(y_t \mid y_{t-1}) &= \mathcal{N}\big(y_t; \sqrt{1-\beta_t}\, y_{t-1}, \, \beta_t \mathbf{I}\big),
\vspace{-0.4cm}
\end{align}
where $\beta_t \in (0,1)$ is a pre-defined noise schedule. The joint forward distribution over all diffusion timesteps $t=1,\dots,T$ is therefore:
\vspace{-0.3cm}
\begin{equation}
q(x_{1:T}, y_{1:T} \mid x_0, y_0) = \prod_{t=1}^T q(x_t \mid x_{t-1}) \, q(y_t \mid y_{t-1}).
\vspace{-0.2cm}
\end{equation}
The closed-form marginal distributions conditioned on the clean data are:
\vspace{-0.2cm}
\begin{align}
q(x_t \mid x_0) &= \mathcal{N}\big(x_t; \sqrt{\bar{\alpha}_t}\, x_0, \, (1-\bar{\alpha}_t)\mathbf{I}\big),\\
q(y_t \mid y_0) &= \mathcal{N}\big(y_t; \sqrt{\bar{\alpha}_t}\, y_0, \, (1-\bar{\alpha}_t)\mathbf{I}\big),
\end{align}
where $\bar{\alpha}_t = \prod_{s=1}^t \alpha_s$ and $\alpha_t = 1-\beta_t$.

\vspace{0.3cm}
\noindent \textbf{Reverse Sampling.}
We consider \textit{(i)}~unconditional generation~(Fig.~\ref{fig:framework}~\textcolor{colora}{(top-right)}), which produces a paired lesion mask and lesion-embedded brain image from Gaussian noise; and \textit{(ii)}~conditional generation~(Fig.~\ref{fig:framework}~\textcolor{colorb}{(bottom-left)}), which synthesizes a brain image given a lesion mask.

\vspace{0.1cm}
\noindent \textbf{\textit{- Unconditional generation:}}\label{parag:unconditional} the joint reverse diffusion process predicts the lesion-brain pair $(x_{t-1}, y_{t-1})$ at each step, conditioned on the current noisy pair $(x_t, y_t)$. To improve anatomical and lesion consistency, we further condition each step on the predicted clean estimates $\hat x_0^t$ and $\hat y_0^t$, which are obtained from a \textbf{\textit{one-step denoising}} of $(x_t, y_t)$. This leads to the modified joint reverse distribution:
\vspace{-0.2cm} 
\begin{equation}
p_\theta(x_{t-1}, y_{t-1} \mid x_t, y_t, \hat x_0^t, \hat y_0^t),
\vspace{-0.2cm}
\label{eq:onestep_denoising}
\end{equation}
where $\hat x_0^t$ guides the generation of lesion masks and $\hat y_0^t$ guides the image generation. Using the chain rule, the joint reverse process can be factorized as:
\vspace{-0.2cm} 
\begin{align}
&p_\theta(x_{t-1}, y_{t-1} \mid x_t, y_t, \hat x_0^t, \hat y_0^t) \numberthis  \\
&= p^{x}_\theta(x_{t-1} \mid x_t, \hat y_0^t) \, p^{y}_\theta(y_{t-1} \mid y_t, \hat x_0^t), \nonumber
\vspace{-0.2cm} 
\end{align}
where $p_\theta^{x}$ corresponds to the \texttt{USB}\textsubscript{lesion} sub-model, $p_\theta^{y}$ corresponds to the \texttt{USB}\textsubscript{brain} sub-model. 
Extending this across all timesteps yields the full joint reverse distribution:
\vspace{-0.2cm} 
\begin{align}
&p_\theta(x_{0:T}, y_{0:T} \mid \hat x_0^t, \hat y_0^t) \numberthis \\ 
&= p(x_T)\,p(y_T) \times \prod_{t=1}^T p^x_\theta(x_{t-1} \mid x_t, \hat y_0^t) \, p^y_\theta(y_{t-1} \mid y_t, \hat x_0^t), \nonumber
\end{align}
where $p(x_T)$ and $p(y_T)$ are isotropic Gaussian priors.
In practice, to avoid circular dependencies, we approximate $p^x_\theta(x_{t-1} \mid x_t, \hat y_0^t)$ by $p^x_\theta(x_{t-1} \mid x_t, y_t)$.

\vspace{0.1cm}
\noindent \textbf{\textit{- Conditional generation:}} a lesion mask $x_0$ is given as an input condition, based on which the corresponding lesion-embedded brain image $y_0$ is generated. Specifically, we sample the brain image sequence starting from noise $y_T$, conditioned on $x_0$, using the \texttt{USB}\textsubscript{brain} model $p_\theta^y$:
\vspace{-0.23cm} 
\begin{equation}
p_\theta(y_{0:T} \mid x_0) = p(y_T) \prod_{t=1}^T p^y_\theta(y_{t-1} \mid y_t, x_0),
\vspace{-0.18cm} 
\end{equation}
$p(y_T)$ is the Gaussian prior at $T$, allowing the model to synthesize a brain image embedded with specific lesion masks.

\vspace{0.1cm}
\noindent \textbf{Training Objective.}
The simplified noise prediction loss can be decomposed into two components, corresponding to the two \texttt{USB} sub-models:
\vspace{-0.1cm} 
\begin{align}
\mathcal{L}_{\text{lesion}} &= \mathbb{E}_{x_0, y_0, \epsilon_x, t} 
\Big[ \|\epsilon_x - \epsilon_\theta^x(x_t, \hat y_0, t)\|^2 \Big], \\
\mathcal{L}_{\text{brain}} &= \mathbb{E}_{x_0, y_0, \epsilon_y, t} 
\Big[ \|\epsilon_y - \epsilon_\theta^y(y_t, \hat x_0, t)\|^2 \Big],
\vspace{-0.55cm} 
\end{align}
where $\epsilon_x$, $\epsilon_y$ and $\epsilon_\theta^x$, $\epsilon_\theta^y$ are the ground truth and predicted noise, respectively. 
The total loss becomes
$
\mathcal{L} = \mathcal{L}_{\text{lesion}} + \mathcal{L}_{\text{brain}},
$
where $\mathcal{L}_{\text{lesion}}$ corresponds to the \texttt{USB}\textsubscript{lesion} sub-model that predicts lesion masks conditioned on the image estimate $\hat x_0$, $\mathcal{L}_{\text{brain}}$ corresponds to the \texttt{USB}\textsubscript{brain} sub-model that predicts brain image conditioned on the lesion estimate $\hat y_0$.

\begin{figure*}
\centering
\includegraphics[width=0.95\linewidth]{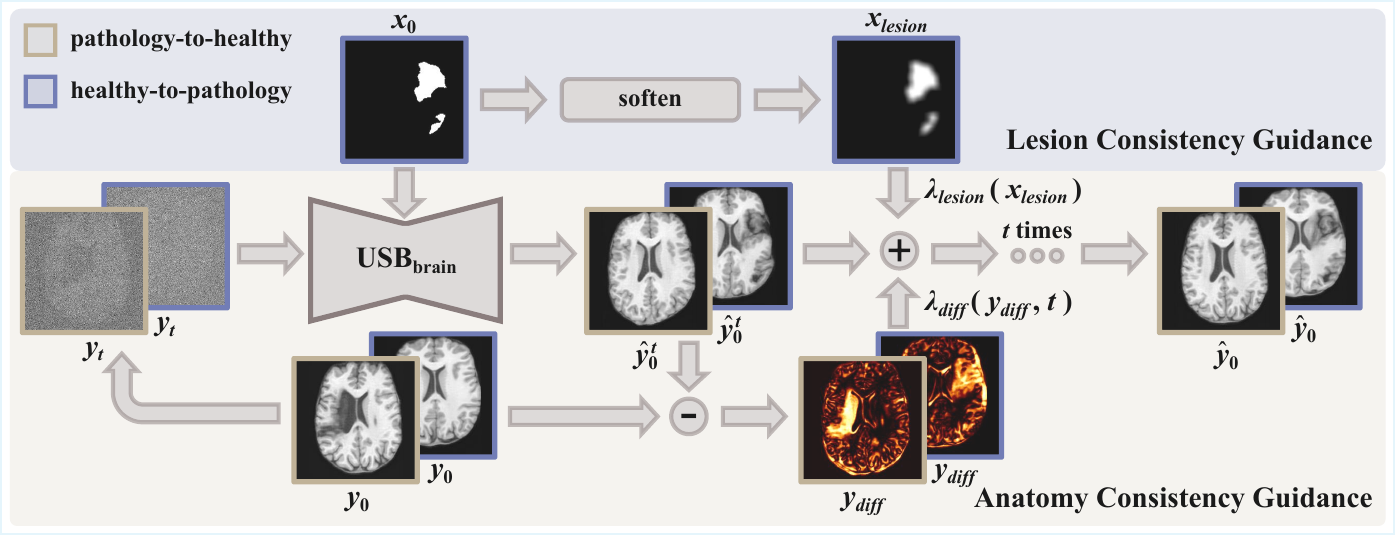}
\vspace{-0.2cm}
\caption{
\textbf{Consistency Guidance Algorithm during editing.} 
\textcolor{colord}{(top)} Lesion Consistency Guidance applied in the healthy-to-pathology editing process.  
\textcolor{colorc}{(bottom)} Anatomy Consistency Guidance applied during both pathology-healthy and healthy-to-pathology editing.}
\label{fig:guidance} 
\vspace{-0.3cm}
\end{figure*}

\subsection{Consistency Guided Bidirectional Editing}
\label{subsec:editing}

We consider two complementary editing directions: \textit{(i)}~pathology-to-healthy editing (Fig.~\ref{fig:framework}~\textcolor{colorc}{(bottom-middle)}) functions as a restorative process that removes existing lesions and reconstructs plausible healthy tissue instead; 
\textit{(ii)}~healthy-to-pathology editing (Fig.~\ref{fig:framework}~\textcolor{colord}{(bottom-right)}) performs a controllable synthesis of pathological brains by embedding specific lesion masks into healthy anatomy.
We formulate \textit{(i)-(ii)} jointly as a projection from a paired input $(x_0, y_0)$ (a lesion mask and its corresponding pathological or healthy brain image) to a target brain $\hat{y}_0$, guided by our Consistency Guidance algorithm which enforces coherence at both global anatomical and regional lesion levels: 

\vspace{0.1cm} 
\noindent \textbf{\textit{- Healthy-to-pathology editing:}} from an originally healthy brain image $y_0$ and a binary lesion mask $x_0$, our objective is to embed a plausible lesion pattern within the healthy brain anatomy, to synthesize a realistic pathological brain image $\hat{y}_0$. 
We first add noise at a randomly chosen timestep $t$ to obtain a noisy latent $y_t$, and then iteratively sample $y_{t-1}$ through the reverse process conditioned on $x_0$.

\vspace{0.1cm} 
\noindent \textbf{\textit{- Pathology-to-healthy editing:}} given a pathological brain image $y_0$, our goal is to reconstruct its healthy counterpart $\hat{y}_0$. Similar to the healthy-to-pathology editing, we first perturb $y_0$ to obtain a noisy representation $y_t$.
The subsequent reverse diffusion process then follows a similar probabilistic formulation as in the conditional generation setting:
$
p^y_\theta(y_{t-1} \mid y_t, x_\varnothing),
$
where $x_\varnothing$ denotes an empty conditional lesion mask, indicating the output brain image $\hat{y}_0$ should be healthy, without any pathological patterns.

%%%%5
%p2h
%%%%%

\vspace{0.1cm} 
{\noindent \textbf{Anatomy Consistency Guidance.}} \label{parag:acg} To preserve anatomical fidelity during both \textbf{\textit{healthy-to-pathology}} and \textbf{\textit{pathology-to-healthy}} editings, we introduce an Anatomy Consistency Guidance algorithm (Fig.~\ref{fig:guidance}~\textcolor{colorc}{(bottom)}).
{Similar to unconditional generation in Sec.~\ref{subsec:generation}, we resort to the one-step denoised image at timestep $t$ (i.e., $\hat y_0^t$), to serve as the current estimation of the underlying healthy anatomy.}
Specifically, we define a \textit{spatiotemporal consistency weight}, $\lambda_{\text{diff}}$, based on the absolute structural difference between the input and the denoised estimate:
\vspace{-0.25cm} 
\begin{equation}
\label{eq:acg}
\lambda_{\text{diff}} = \exp\big(-\alpha_t \cdot y_{\text{diff}}\big), \quad \alpha_t = \alpha_0 \cdot \exp\left(-k \frac{t}{T}\right),
\vspace{-0.15cm} 
\end{equation}
where $y_{\text{diff}} = |y_0 - \hat y_0^t|$, $\alpha_0$ controls the initial guidance strength, and $k$ modulates its temporal attenuation across timesteps. 
\textit{(i)~Spatially}, the pixel-wise modulation by $y_{\text{diff}}$ enables an adaptive consistency: regions with large $y_{\text{diff}}$ correspond to lesion areas, leading to smaller $\lambda_{\text{diff}}$ which in turn offers the model greater editing flexibility within pathological regions; in contrast, regions with small $y_{\text{diff}}$ correspond to healthy tissue, where larger $\lambda_{\text{diff}}$ enforces stronger structural preservation. 
\textit{(ii)~Temporally}, the exponential decay term $\alpha_t$ provides dynamic control over the strength of guidance: at early timesteps, smaller $\alpha_t$ values grant the model greater freedom to remove lesions and perform aggressive restoration, whereas at later timesteps, the increasing $\alpha_t$ amplifies $\lambda_{\text{diff}}$, constraining the model to preserve anatomical consistency in the final reconstruction.

Finally, we update the reverse latent for both editing as:
\vspace{-0.2cm} 
\begin{equation}
y_{t-1} \leftarrow y_{t-1} + \lambda_{\text{diff}} \cdot (y_0 - y_{t-1}).
\vspace{-0.2cm} 
\end{equation}
This adaptively preserves the anatomical structure of the original brain while encouraging the removal of pathology.  
The updated $y_{t-1}$ is then used for the next reverse diffusion step, ensuring consistent and anatomically coherent editing.

%%%%5
%h2p
%%%%%

\vspace{0.1cm} 
{\noindent \textbf{Lesion Consistency Guidance.}} For \textbf{\textit{healthy-to-pathology}} editing, we further propose a Lesion Consistency Guidance algorithm (Fig.~\ref{fig:guidance}~\textcolor{colord}{(top)}), to enhance the localization and morphological embedding of lesion patterns. Specifically, we define a \textit{lesion-guided weight} based on a spatially smooth prior (a soft lesion map through average pooling):
\vspace{-0.2cm} 
\begin{equation}
\label{eq:lcg}
\lambda_{\text{lesion}} = 1 - \eta \cdot x_{\text{lesion}}, \quad x_{\text{lesion}} = \operatorname{AvgPool}(x_0),
\vspace{-0.2cm} 
\end{equation}
where $\eta$ controls the degree of lesion-free scaling, suppressing excessive or unnecessary edits outside lesion regions. The final adaptive consistency weight becomes 
$\lambda = \lambda_{\text{diff}} \cdot \lambda_{\text{lesion}},
$
forming a guided update for each reverse step:
\vspace{-0.2cm} 
\begin{equation}
y_{t-1} \leftarrow y_{t-1} + \lambda \cdot (y_0 - y_{t-1}),
\vspace{-0.2cm} 
\end{equation}
where the difference term $(y_0 - y_{t-1})$ drives the edited image toward the original anatomy while allowing lesion-specific deviations modulated by $\lambda$. The updated $y_{t-1}$ is then propagated to the next reverse diffusion step.

%%%%%%%%%%%%%%%%%%%%%%%%%%%%%
\input{fig_tex/fig_generation} 
%%%%%%%%%%%%%%%%%%%%%%%%%%%%%

\section{Experiments}
\label{sec:exp}

\vspace{0.1cm}
\noindent \textbf{Datasets.} We conducted experiments on \textit{six} public brain MRI datasets (ADNI~\cite{Jack2008TheAD}, HCP~\cite{Essen2012TheHC}, ADHD200~\cite{Brown2012ADHD200GC}, OASIS3~\cite{LaMontagne2018OASIS3LN}, ATLAS~\cite{Liew2017ATLAS}, and ISLES~\cite{Hernandez2022ISLES}), spanning healthy controls, stroke patients, and Alzheimer’s disease cases with white matter hyperintensity (WMH). All images were skull-stripped~\cite{Hoopes2022SynthStripSF} and resampled to an isotropic voxel spacing of $1~\mathrm{mm}$ for consistency. 
ATLAS ($n=655$ T1-weighted (T1w) MRIs) and ISLES ($n=152$ FLAIR MRIs) are stroke cohorts, provided with gold-standard manual lesion segmentations.
ADNI ($n=2045$ T1w MRIs) includes Alzheimer’s patients with WMH, for which WMH labels were obtained with LST-AI~\cite{wiltgen2024lst}.
The remaining datasets (HCP ($n=897$ T1w MRIs), ADHD200 ($n=961$ T1w MRIs), and OASIS3 ($n=883$ T1w MRIs)), consist of healthy subjects with normal brain anatomy.

\vspace{0.1cm}
\noindent \textbf{Evaluation Metrics.} For generation, we use FID, KID, and MMD, following the synthetic image evaluation protocol in ~\cite{sun2022hierarchical}. For editing, we use L1 distance, PSNR, and SSIM.

\vspace{0.1cm}
\noindent \textbf{Implementation Details.}
{As a general framework for generation-editing, \texttt{USB} can use any diffusion-based backbone.}
We adopt the standard 3D Latent Diffusion Model (LDM)~\cite{rombach2022ldm}. The network follows a 3D U-Net architecture, its autoencoder employs a KL-regularized latent space with perceptual and patch-based adversarial objectives to map 3D brain images into a compact latent representation.

For stroke, we generate pseudo-pathological training samples using an anomaly randomization strategy~\cite{liu2025una} that augments and diversifies real lesion masks drawn from the ATLAS and ISLES datasets. 
These randomized pseudo-lesions are then used as conditioning lesion masks and embedded into healthy brain images from HCP, ADHD200, and OASIS3, which serve as ground-truth healthy counterparts for quantitative evaluation.
For WMH, we use real pathological images from ADNI dataset as training samples.
All training volumes are resized at $160^3$. The noise schedule $\beta_t$ increases linearly from $\beta_1 = 10^{-4}$ to $\beta_T = 0.02$, with a maximum diffusion step $T = 1024$. During inference, we sample with $T = 300$ steps for efficiency. For the consistency guidance mechanisms, we set $\alpha_0 = 20$ and $k = 0.5$ for Anatomy Consistency Guidance (ACG), and $\eta = 1$ for Lesion Consistency Guidance (LCG). All experiments are conducted on a single NVIDIA A40 GPU.

\vspace{0.1cm}
\noindent \textbf{Competing Models.}
\texttt{USB} is the first unified model that simultaneously achieves bidirectional generation and editing between pathological and healthy brain images, thereby establishing the first benchmark on unified task evaluation. 
In addition, we include the following state-of-the-art methods for comparison, which are most relevant to the individual tasks of pathology-to-healthy and healthy-to-pathology synthesis:  
\textit{(i)}~\texttt{SynthSR}~\cite{iglesias2023synthsr} (2023), a super-resolution model for T1w brain MRI synthesis;  
\textit{(ii)}~\texttt{Brain-ID}~\cite{liu2024brainid} (2024), a generative model for brain anatomy representation; and  
\textit{(iii)}~\texttt{UNA}~\cite{liu2025una} (2025), a model for pathology-to-healthy image translation, trained on samples synthesized via a complementary pathology-encoding strategy~\cite{liu2024pepsi}. 
\texttt{SynthSR} and \texttt{Brain-ID} are designed for healthy image synthesis and \textit{cannot} perform healthy-to-pathology editing.

%%%%%%%%%%%%%%%%%%%%%%%%%%%%%
\input{fig_tex/fig_editing_comparison}
%%%%%%%%%%%%%%%%%%%%%%%%%%%%%

\subsection{Bidirectional Pathology-Healthy Generation}
\label{subsec:generation_exp}

%%%%%%%%%%%%%%%%%%%%%%%%%%%%%
\begin{table}[t]
\centering
{\fontsize{6pt}{6pt}\selectfont
\caption{\textbf{Quantitative results of (un)conditional generation.}}
\vspace{-0.2cm} %调整图片与上文的垂直距离
\label{tab:generation}

\resizebox{\linewidth}{!}{
\begin{tabular}{clccc}
\toprule
{\textbf{Lesion}} & \textbf{Task} & \textbf{FID} ($\downarrow$) & \textbf{KID} ($\downarrow$) & \textbf{MMD} ($\downarrow$) \\ 
\midrule

% ===== stroke =====
\multirow{2}{*}{{\texttt{Stroke}}} 
& \texttt{unconditional}  & 0.0073 & 0.0018 & 0.0148 \\
& \texttt{conditional} & 0.0109 & 0.0037 & 0.0170  \\
\cmidrule(lr){1-5}

% ===== WMH =====
\multirow{2}{*}{{\texttt{WMH}}}

& \texttt{unconditional} & 0.0338 & 0.0223 & 0.1084 \\
& \texttt{conditional} & 0.0346 & 0.0177 & 0.0947 \\
\bottomrule
\end{tabular}
}}
\vspace{-0.3cm}
\end{table}

%%%%%%%%%%%%%%%%%%%%%%%%%%%%%

We first evaluate \texttt{USB} and establish the first benchmark on two generation tasks in our unified framework (Fig.~\ref{fig:task}): \textbf{\textcolor{colora}{(a)}}~unconditional paired lesion–brain image generation; \textbf{\textcolor{colorb}{(b)}}~conditional brain image generation given a lesion mask.

\vspace{0.1cm}
\noindent \textbf{Unconditional Generation.}
Fig.~\ref{fig:generation}~(a) presents examples of paired lesion masks and brain images generated by \texttt{USB}. 
Notably, although the paired lesion-brain generation begins \textit{unconditionally} from random noise, \texttt{USB} produces lesion masks exhibiting diverse locations, sizes, and shapes, while simultaneously generating anatomically coherent brain images with embedded pathological patterns that \textit{spatially correspond} to the generated lesions within each pair. 

\vspace{0.1cm}
\noindent \textbf{Conditional Generation.}
Fig.~\ref{fig:generation}~(b) shows conditional generation results given a lesion mask. Conditioned on a single lesion mask, \texttt{USB} can generate an \textit{unlimited} brain image samples that embed the specified lesion, producing diverse anatomical appearances while maintaining faithful pathological localization. As illustrated by the samples along each row, the generated pathological images exhibit variations in intensity, texture, and local tissue response.

In Tab.~\ref{tab:generation}, we present the first quantitative benchmark evaluation of both generation tasks.  
For the unconditional generation task, we generated 256 paired lesion–brain samples for each lesion type (stroke and WMH).  
For the conditional generation task, we used masks from the ATLAS and ISLES test sets as stroke lesions, and masks from the ADNI test set as WMH lesions.  
Notably, for each lesion type, the unconditional generation achieved superior quantitative performance, highlighting that the paired diffusion mechanism effectively generates coherent lesion-brain pairs through joint modeling between \texttt{USB}\textsubscript{brain} and \texttt{USB}\textsubscript{lesion}. 
{Additional generation results are provided in Suppl.~\ref{suppl:generation_exp}.}

These results demonstrate \texttt{USB}’s ability to generate diverse yet anatomically plausible pathological patterns, thereby enabling the creation of large-scale brain imaging datasets with paired lesion annotations, effectively bypassing the need for labor-intensive manual labeling.

%%%%%%%%%%%%%%%%%%%%%%%%%%%%%
%%%%%%%%%%%%%%%%%%%%%%%%%%%%%

%%%%%%%%%%%%%%%%%%%%%%%%%%%%%
\begin{table}[t]
\centering
\caption{\textbf{Quantitative results of pathology-to-healthy editing.}}
\vspace{-0.2cm}
\label{tab:editing_p2h}
{\fontsize{8pt}{8pt}\selectfont
\resizebox{\linewidth}{!}{
\begin{tabular}{clccc}
\toprule
\thead{\textbf{Dataset}\\(Train/Test)} & \textbf{Method} & \textbf{L1} ($\downarrow$) & \textbf{PSNR} ($\uparrow$) & \textbf{SSIM} ($\uparrow$) \\ 
\midrule

% ===== HCP =====
\multirow{4}{*}{\thead{\texttt{HCP}~\cite{Essen2012TheHC}\\(808/89)}} 
& \texttt{SynthSR}~\cite{iglesias2023synthsr} & 0.048 & 21.37 & 0.831 \\
& \texttt{Brain-ID}~\cite{liu2024brainid} & 0.077 & 27.47 & 0.844 \\
& \texttt{UNA}~\cite{liu2025una} & 0.041 & 25.56 & 0.959 \\
& \texttt{USB} & \textbf{0.009} & \textbf{32.31} & \textbf{0.967} \\
\cmidrule(lr){1-5}

% ===== ADHD200 =====
\multirow{4}{*}{\thead{\texttt{ADHD200}~\cite{Brown2012ADHD200GC}\\(865/96)}}
& \texttt{SynthSR}~\cite{iglesias2023synthsr} & 0.092 & 16.69 & 0.737 \\
& \texttt{Brain-ID}~\cite{liu2024brainid} & 0.119 & 29.71 & 0.781 \\
& \texttt{UNA}~\cite{liu2025una} & 0.018 & 29.71 & \textbf{0.974} \\
& \texttt{USB} & \textbf{0.008} & \textbf{33.25} & 0.968 \\
\cmidrule(lr){1-5}

% ===== OASIS3 =====
\multirow{4}{*}{\thead{\texttt{OASIS3}~\cite{LaMontagne2018OASIS3LN}\\(601/67)}}
& \texttt{SynthSR}~\cite{iglesias2023synthsr} & 0.082 & 16.78 & 0.733 \\
& \texttt{Brain-ID}~\cite{liu2024brainid} & 0.077 & 17.17 & 0.830 \\
& \texttt{UNA}~\cite{liu2025una} & 0.016 & 29.49 & 0.978 \\
& \texttt{USB} & \textbf{0.007} & \textbf{33.56} & \textbf{0.980} \\

\bottomrule
\end{tabular}
}}
\vspace{-0.3cm}
\end{table}
%%%%%%%%%%%%%%%%%%%%%%%%%%%%%

%%%%%%%%%%%%%%%%%%%%%%%%%%%%%
% \input{tab_tex/tab_h2p}
%%%%%%%%%%%%%%%%%%%%%%%%%%%%%

\subsection{Bidirectional Pathology-Healthy Editing}
\label{subsec:editing_exp}

In addition to generation, we further evaluate \texttt{USB} and set up the first benchmark on the two editing tasks in the unified framework (as shown in Fig.~\ref{fig:task}): \textbf{\textcolor{colorc}{(c)}}~pathology-to-healthy editing, and \textbf{\textcolor{colord}{(d)}}~healthy-to-pathology editing. 
To enable quantitative evaluation of pathology-to-healthy editing, we randomly synthesize pathological inputs from healthy images for testing, and use the originals as ground truth.

%%%%%%%%%%%%%%%%%%%%%%%%%%%%%
%\input{sec/X_suppl_editing}
%%%%%%%%%%%%%%%%%%%%%%%%%%%%%

\vspace{0.1cm}
\noindent \textbf{Pathology-to-Healthy Editing.}
Fig.~\ref{fig:editing_compare}~(a) presents comparison results on four input brain images with lesions of varying sizes, shapes, and densities.  
All competing models, although specialized for healthy brain image synthesis, struggle to reconstruct pathological regions, especially near lesion boundaries. 
In contrast, \texttt{USB} accurately reconstructs the corresponding healthy brain, even in challenging cases with large, high-density lesions where normal brain structures are almost \textit{completely obscured} (last row).
Tab.~\ref{tab:editing_p2h} provides a quantitative comparison for pathology-to-healthy editing, where \texttt{USB} achieves the best performance across all three datasets, demonstrating the effectiveness of its unified framework in reconstructing healthy brains from pathological inputs. 
{Additional results are in Suppl.~\ref{suppl:editing_exp}.}

\vspace{0.1cm}
\noindent \textbf{Healthy-to-Pathology Editing.}
As shown in Fig.~\ref{fig:editing_compare}~(b), given a healthy brain image paired with a random lesion mask, \texttt{USB} seamlessly embeds the lesion into the healthy anatomy, producing realistic pathological appearances consistent with surrounding structural context. Yet \texttt{UNA}'s syntheses are \textit{visually unrealistic}, with the conditioned lesion mask simply overlaid as a parallel layer onto the healthy brain. 
Tab.~\ref{tab:editing_h2p} further demonstrates \texttt{USB}'s superior performance. The evaluation was conducted on 100 pairs of lesion masks and healthy brains, with real stroke images\begin{wraptable}{r}{0.25\textwidth}
\vspace{-10pt}
\begin{adjustwidth}{-0.5cm}{}
\setlength{\intextsep}{0pt}
\setlength{\columnsep}{-5pt}
\raggedleft
\centering
\renewcommand{\arraystretch}{1}
\scriptsize
\caption{\textbf{Quantitative results of healthy-to-pathology editing.}}
\label{tab:editing_h2p}
\vspace{-0.2cm}
\setlength{\tabcolsep}{6pt}
\begin{tabular}{@{}lccc@{}}
\toprule
\textbf{Method} & \textbf{FID} ($\downarrow$) & \textbf{KID} ($\downarrow$) & \textbf{MMD} ($\downarrow$) \\ 
\midrule
\texttt{UNA}~\cite{liu2025una} & 0.4017 & 0.2752 & 0.6012 \\
\texttt{USB} & \textbf{0.2533} & \textbf{0.1949} & \textbf{0.6006} \\
\bottomrule
\end{tabular}
\vspace{-0.5cm}
\end{adjustwidth}
\end{wraptable}from ATLAS dataset used as reference for FID, KID, and MMD. \texttt{USB} consistently outperforms \texttt{UNA} across all metrics.

%%%%%%%%%%%%%%%%%%%%%%%%%%%%%
%%%%%%%%%%%%%%%%%%%%%%%%%%%%%

\subsection{Ablation Studies}
\label{subsec:ablation}

%%%%%%%%%%%%%%%%%%%%%%%%%%%%%
\input{fig_tex/fig_ablat_acg} 
%%%%%%%%%%%%%%%%%%%%%%%%%%%%%

In this section, we conduct ablation studies on the proposed Consistency Guidance mechanisms within the editing tasks (Sec.~\ref{parag:acg}). 
Additional ablations on one-step denoising condition~(Sec~\ref{parag:unconditional}) and model hyperparameters are in Suppl.~\ref{suppl:ablation_exp}.

\vspace{0.1cm}
\noindent \textbf{Anomaly Consistency Guidance.}
Fig.~\ref{fig:ablation_acg} demonstrates the effectiveness of the proposed Anatomy Consistency Guidance (ACG) on bidirectional editing. 
Without ACG, the generated brains exhibit noticeable and unrealistic anatomical deformations in non-lesion regions (arrow-indicated), leading to structural inconsistencies between the edited and original brains, as shown in the corresponding difference maps. After incorporating ACG, the model effectively constrains editable modifications \textit{within} pathological regions, while preserving the remaining anatomy. 
Tab.~\ref{tab:ablation_acg} further reports their quantitative performances: removing ACG leads to a substantial degradation in all L1, PSNR, and SSIM scores, reaffirming the critical role of ACG in achieving more accurate and structurally consistent reconstructions.

%%%%%%%%%%%%%%%%%%%%%%%%%%%%%
\begin{table}[t]
%\vspace{-0.5cm} 
\centering
{\fontsize{7pt}{8pt}\selectfont
\caption{\textbf{Quantitative result of ablation study on Anomaly Consistency Guidance (ACG) for pathology-to-healthy editing.}}
\vspace{-0.2cm} 
\label{tab:ablation_acg}
\resizebox{\linewidth}{!}{
\begin{tabular}{clccc}
\toprule
\textbf{Dataset} & \textbf{Method} & \textbf{L1} ($\downarrow$) & \textbf{PSNR} ($\uparrow$) & \textbf{SSIM} ($\uparrow$) \\ 
\midrule

% ===== HCP =====
\multirow{2}{*}{\thead{\texttt{HCP}~\cite{Essen2012TheHC}}} 
& \text{\textit{\textit{w/}o} \texttt{ACG}} & 0.025 & 25.56 & 0.835 \\
& \text{\textit{w/{\color{white}o}} \texttt{ACG}} & \textbf{0.009} & \textbf{32.31} & \textbf{0.967} \\
\cmidrule(lr){1-5}

% ===== ADHD200 =====
\multirow{2}{*}{\thead{\texttt{ADHD200}~\cite{Brown2012ADHD200GC}}}
& \text{\textit{\textit{w/}o} \texttt{ACG}} & 0.238 & 26.14 & 0.826 \\
& \text{\textit{w/{\color{white}o}} \texttt{ACG}} & \textbf{0.008} & \textbf{33.25} & \textbf{0.968} \\
\cmidrule(lr){1-5}

% ===== OASIS3 =====
\multirow{2}{*}{\thead{\texttt{OASIS3}~\cite{LaMontagne2018OASIS3LN}}}
& \text{\textit{\textit{w/}o} \texttt{ACG}} & 0.026 & 25.10 & 0.849 \\
& \text{\textit{w/{\color{white}o}} \texttt{ACG}} & \textbf{0.007} & \textbf{33.56} & \textbf{0.980} \\

\bottomrule
\end{tabular}
}}
\vspace{-0.3cm}
\end{table} 
%%%%%%%%%%%%%%%%%%%%%%%%%%%%%

%%%%%%%%%%%%%%%%%%%%%%%%%%%%%
\input{fig_tex/fig_ablat_lcg} 
%%%%%%%%%%%%%%%%%%%%%%%%%%%%%

\vspace{0.1cm}
\noindent \textbf{Lesion Consistency Guidance.}
Fig.~\ref{fig:ablation_lcg} illustrates the impact of Lesion Consistency Guidance (LCG) for healthy-to-pathology editing. Without LCG, the generated lesion regions appear blurry around lesion boundaries, and the pathological patterns are less distinct, resulting in ambiguous lesion localization and weak pathological expression. Incorporating LCG produces lesions with more precise localization and well-defined shapes that better align with the given lesion masks, as well as textures and contrasts that exhibit stronger pathological characteristics. These results demonstrate that LCG effectively enforces lesion-aware conditioning, enhancing both spatial and semantic consistency of the generated pathological regions.

\section{Future Work and Broader Impact}
\label{sec:future}

%\noindent \textbf{Limitations.}

% \vspace{0.1cm}
\noindent \textbf{Scalable Dataset Creation.} 
For the first time, \texttt{USB} enables the generation of \textit{bidirectional, paired} pathological–healthy datasets with ground-truth lesion masks, providing a scalable alternative to manual segmentation and supporting supervised pathology–healthy translation. These paired syntheses can also simulate disease progression and recovery, supporting clinical visualization and treatment planning.

\vspace{0.1cm}
\noindent \textbf{Robust Neuroimaging Analysis.} Conventional pipelines (e.g., FreeSurfer~\cite{fischl2002freesurfer}, NiftyReg~\cite{modat2010fast}, ANTs~\cite{avants2009ants}) often fail on scans with severe lesions or missing tissue. \texttt{USB}’s pathology-to-healthy editing recovers healthy anatomy, improving the robustness of existing neuroimaging tools, and enhancing downstream morphometric analysis. 

We plan to extend \texttt{USB} to cross-modality generation and editing and integrate it into routine neuroimaging workflows, to assess its impact on diverse clinical applications.

\section{Conclusion}
\label{sec:conclusion}
We introduce \texttt{USB}, the first unified framework for bidirectional brain image generation and editing that jointly models lesion and brain anatomy. 
Our paired diffusion mechanism ensures anatomically coherent generation for both pathological and healthy images, while the proposed consistency guidance preserves original brain anatomy during lesion editing.
\texttt{USB} supports unconditional paired lesion–brain generation, conditional generation from lesion masks, and bidirectional pathology-healthy editing --- all within a single framework and without task-specific retraining.
We demonstrate \texttt{USB}'s superior performance on six public brain image datasets spanning diverse patient cohorts.
By unifying generation and editing across pathological and healthy domains, we believe \texttt{USB} establishes the foundation for scalable and robust neuroimaging analysis. \par\nobreak
{
    \small 
    \bibliographystyle{ieeenat_fullname}
    \bibliography{main}
}


\clearpage
\setcounter{figure}{0}
\setcounter{equation}{0}
\setcounter{section}{0}

\renewcommand{\thesection}{\Alph{section}}
\renewcommand\thefigure{\thesection.\arabic{figure}}
\renewcommand\thefigure{\thesection.\arabic{table}}
\renewcommand\theequation{\thesection.\arabic{equation}}
\counterwithin{figure}{section}
\counterwithin{table}{section}
\counterwithin{equation}{section}

\maketitlesupplementary

%\section{Rationale}
%\label{sec:rationale}
% 
\noindent This supplementary material includes: 
\begin{itemize}[leftmargin=-1pt]

\item[] {\bf{\ref{suppl:generation_exp}:}} Experimental results on bidirectional pathology-healthy \textit{generation}, complementing Sec.~\ref{subsec:generation_exp}.

\item[] {\bf{\ref{suppl:editing_exp}:}} Experimental results on bidirectional pathology-healthy \textit{editing}, complementing Sec.~\ref{subsec:editing_exp}. 

\item[] {\bf{\ref{suppl:ablation_exp}:}} Additional ablation studies on one-step denoising conditioning~(Sec~\ref{parag:unconditional}) and hyperparameter selections of the proposed anatomy and lesion consistency guidance algorithms, complementing Sec.~\ref{subsec:ablation}.

\item[] {\bf{\ref{suppl:exp_real}:}} Further experiments and discussion on \texttt{USB}’s effectiveness and broader impact in challenging clinical scenarios, including healthy tissue reconstruction for patient scans, and image artifacts removal and inpainting. %, where \textbf{\textit{no ground-truth exists}}.

\item[] {\bf{\ref{suppl:limitation}:}} Discussion on \texttt{USB}'s limitations.

\item[] {\bf{\ref{suppl:dataset}:}} Additional implementation details on datasets and evaluation metrics.

\item[] {\bf{\ref{suppl:preliminary}:}} Preliminary background and notational formulation of diffusion models.

\end{itemize}
% 
%To split the supplementary pages from the main paper, you can use \href{https://support.apple.com/en-ca/guide/preview/prvw11793/mac#:~:text=Delete%20a%20page%20from%20a,or%20choose%20Edit%20%3E%20Delete).}{Preview (on macOS)}, \href{https://www.adobe.com/acrobat/how-to/delete-pages-from-pdf.html#:~:text=Choose%20%E2%80%9CTools%E2%80%9D%20%3E%20%E2%80%9COrganize,or%20pages%20from%20the%20file.}{Adobe Acrobat} (on all OSs), as well as \href{https://superuser.com/questions/517986/is-it-possible-to-delete-some-pages-of-a-pdf-document}{command line tools}.

%%%%%%%%%%%%%%%%%%%%%%%%%%%%%
%%%%%%%%%%%%%%%%%%%%%%%%%%%%%
%%%%%%%%%%%%%%%%%%%%%%%%%%%%%

\section{Bidirectional Pathology-Healthy Generation}
\label{suppl:generation_exp}

\input{fig_tex/fig_generation_wmh}

Figs.~\ref{fig:generation_wmh_uncond}-\ref{fig:generation_wmh_cond} illustrate the generation results on the white matter hyperintensity (WMH) dataset. 
Specifically, Fig.~\ref{fig:generation_wmh_uncond} shows unconditional generation of lesion mask and corresponding brain image pairs, 
while Fig.~\ref{fig:generation_wmh_cond} demonstrates conditional generation, where a given lesion mask is used to produce multiple brain images with the lesion embedded.
Compared to the stroke dataset, lesions in the WMH dataset exhibit more consistent shapes and less variability in size, location, and morphology. 
Nevertheless, \texttt{USB} is still able to generate WMH lesion masks and corresponding brain images that display noticeable diversity, capturing subtle variations in lesion appearance while preserving anatomical plausibility.

% Additional qualitative results from Sec.~\ref{subsec:generation_exp}.

%%%%%%%%%%%%%%%%%%%%%%%%%%%%%
%%%%%%%%%%%%%%%%%%%%%%%%%%%%%
%%%%%%%%%%%%%%%%%%%%%%%%%%%%%

\section{Bidirectional Pathology-Healthy Editing}
\label{suppl:editing_exp}

\input{fig_tex/fig_editing}

% Additional qualitative results from Sec.~\ref{subsec:editing_exp}.

\vspace{0.1cm}
\noindent \textbf{Pathology-to-Healthy Editing.}
Fig.~\ref{fig:editing_p2h} illustrates the pathology-to-healthy editing results. Each case shows a brain image containing lesions of different sizes, shapes, and textures. \texttt{USB} successfully identifies and restores the lesion regions, reconstructing plausible healthy tissue. Notably, for large and complex lesions (e.g., the last case), \texttt{USB} can still reconstruct coherent healthy anatomy and even recover structural details that were missing within the lesion area, demonstrating its strong capability in anatomy-aware restoration.
% Tab.~\ref{tab:editing_p2h}

\vspace{0.1cm}
\noindent \textbf{Healthy-to-Pathology Editing.}
Fig.~\ref{fig:editing_h2p}  shows the results of healthy-to-pathology editing. We present two examples of healthy brains, each combined with a distinct lesion mask. For the same lesion mask, \texttt{USB} produces pathology-embedded images with lesion regions that adapt naturally to the anatomical context of each healthy brain. Conversely, for the same healthy brain, providing different lesion masks results in pathology localized to the corresponding regions. This demonstrates that \texttt{USB} effectively integrates lesion semantics into brain anatomy while maintaining structural realism and diversity in pathological synthesis.

%%%%%%%%%%%%%%%%%%%%%%%%%%%%%
%%%%%%%%%%%%%%%%%%%%%%%%%%%%%
%%%%%%%%%%%%%%%%%%%%%%%%%%%%%

\section{Additional Ablation Studies}
\label{suppl:ablation_exp}

\subsection{One-Step Denoising}
\label{suppl:onestep_denoising_exp}

\input{fig_tex/fig_wo_onestep_denoising}

In Sec.~\ref{subsec:generation}, we compute one-step denoising estimates for both modalities. 
Rather than treating each modality independently, we obtain modality-specific noise predictions that are conditioned on the current noisy pair $(x_t,y_t)$, and then form deterministic one-step estimates for $\hat x_0^t$ and $\hat y_0^t$:
\begin{align}
&p_\theta(\hat{x}_0^t \mid x_t, y_t) \\
&= \mathcal{N}\big(\hat{x}_0^t;\; \mu_{\theta}^{0,x}(x_t,y_t,t),\;  \Sigma_{\theta}^{0,x}(x_t,y_t,t)\big),  \nonumber
\end{align}
\begin{align}
&p_\theta(\hat{y}_0^t \mid y_t, x_t) \\
&= \mathcal{N}\big(\hat{y}_0^t;\; \mu_{\theta}^{0,y}(y_t,x_t,t),\; \Sigma_{\theta}^{0,y}(y_t,x_t,t)\big),  \nonumber
\end{align}
with deterministic estimates obtained by setting the covariances to zero and using noise-prediction networks $\epsilon_\theta^x$ and $\epsilon_\theta^y$ conditioned on the noisy pair:
\begin{align}
&\mu_{\theta}^{0,x}(x_t,y_t,t)
= \frac{1}{\sqrt{\bar\alpha_t}}\Big(x_t - \sqrt{1-\bar\alpha_t}\,\epsilon_\theta^x(x_t,y_t,t)\Big), \\
&\Sigma_{\theta}^{0,x}(x_t,y_t,t)=0,
\label{eq:onestep_x_joint}
\end{align}
\begin{align}
&\mu_{\theta}^{0,y}(y_t,x_t,t)
= \frac{1}{\sqrt{\bar\alpha_t}}\Big(y_t - \sqrt{1-\bar\alpha_t}\,\epsilon_\theta^y(y_t,x_t,t)\Big), \\
&\Sigma_{\theta}^{0,y}(y_t,x_t,t)=0.
\label{eq:onestep_y_joint}
\end{align}
The resulting pair $(\hat x_0^t,\hat y_0^t)$ is then used as condition for the joint reverse update in Eq.~(\ref{eq:onestep_denoising}).

To verify the effectiveness of conditioning on $(\hat x_0^t,\hat y_0^t)$, we conduct an ablation study. 
As shown in Fig.~\ref{fig:wo_onestep_denoising}, instead of using the one-step denoised estimates $(\hat x_0^t,\hat y_0^t)$, we condition the reverse process directly on the model predictions at timestep $t\!-\!1$, i.e., $(x_{t-1}, y_{t-1})$. 
Compared with the results in Fig.~\ref{fig:generation}~(a), the unconditional generations obtained \emph{without} one-step denoising exhibit noticeably worse anatomical fidelity: the generated brain images show structural inconsistencies, distorted cortical regions, and blurred tissue boundaries. 
This demonstrates that one-step denoising provides a cleaner and more stable anatomical anchor at each diffusion step, thereby improving anatomical consistency across timesteps.

\subsection{ACG and LCG Hyperparameters}
\label{suppl:acg_lcg_exp}

% \todo{Hyperparameters of ACG and LCG}

\input{fig_tex/fig_ablat_hyperparam_acg}

\input{fig_tex/fig_ablat_hyperparam_lcg}

%%%%%%%%%%%%%%%%%%%%%%%%%%%%%
\begin{table}[t]
\centering
\caption{\textbf{Quantitative ablation results for the hyperparameters used in Anatomy Consistency Guidance (ACG).}}
\vspace{-0.2cm}
\label{tab:hyperparam_acg}
{\fontsize{8pt}{8pt}\selectfont
\resizebox{\linewidth}{!}{
\begin{tabular}{clccc}
\toprule
\thead{\textbf{Hyperparamter}} & \textbf{Value} & \textbf{L1} ($\downarrow$) & \textbf{PSNR} ($\uparrow$) & \textbf{SSIM} ($\uparrow$) \\ 
\midrule

\multirow{4}{*}{\thead{\texttt{$\alpha$}}} 
& \texttt{5} & 0.008 & 31.74 & 0.974 \\
& \texttt{10} & 0.007 & 32.57 & 0.976 \\
& \texttt{20}  (\texttt{USB}) & 0.009 & 32.31 & 0.967 \\
& \texttt{30} & 0.015 & 29.33 & 0.931 \\
\cmidrule(lr){1-5}

\multirow{3}{*}{\thead{\texttt{$k$}}} 
& \texttt{0.1} & 0.010 & 31.53 & 0.959 \\
& \texttt{0.5} (\texttt{USB}) & 0.009 & 32.31 & 0.967 \\
& \texttt{1} & 0.007 & 33.00 & 0.973 \\
% & \texttt{30} & 0.015 & 29.33 & 0.931 \\
% \cmidrule(lr){1-5}

\bottomrule
\end{tabular}
}}
\vspace{-0.3cm}
\end{table}
%%%%%%%%%%%%%%%%%%%%%%%%%%%%%

We conduct ablation studies on the hyperparameters of both Anatomy Consistency Guidance (ACG) and Lesion Consistency Guidance (LCG).
For ACG, we systematically vary the parameters $\alpha$ and $k$ in Eq.~(\ref{eq:acg}), while for LCG we evaluate the effect of different values of $\eta$ in Eq.~(\ref{eq:lcg}).

\vspace{0.1cm}
\noindent \textbf{Anatomy Consistency Guidance (ACG).}
Specifically, for ACG, we conduct an ablation study on the pathology-to-healthy editing results using the HCP dataset. 
Figs.~\ref{fig:hyperpara_acg_alpha}-\ref{fig:hyperpara_acg_k} present qualitative results under varying ACG hyperparameters, Tab.~\ref{tab:hyperparam_acg} further reports the corresponding quantitative comparisons. 

As shown in Fig.~\ref{fig:hyperpara_acg_alpha}, a smaller $\alpha$ (e.g., $\alpha=5$) enforces a stronger constraint from the original pathological image, causing residual lesion structures to persist in the edited healthy reconstruction. Increasing $\alpha$ progressively attenuates the influence of the pathological input, leading to more complete lesion removal. At $\alpha=20$, lesions are effectively eliminated and the corresponding healthy brain anatomy is recovered. However, excessively large values of $\alpha$ (e.g., $\alpha=30$) reduce guidance from the original anatomy, and although lesions are removed, subtle geometric deviations may arise in otherwise healthy regions. Quantitative results in Tab.~\ref{tab:hyperparam_acg} indicate that smaller $\alpha$ values achieve higher numerical performance, reflecting stronger anatomical fidelity. Nevertheless, to balance the ultimate objective of fully removing lesions while reconstructing plausible healthy anatomy, we adopt $\alpha=20$ in all experiments, ensuring a trade-off between quantitative accuracy and qualitative editing quality.

Fig.~\ref{fig:hyperpara_acg_k} illustrates the effect of $k$ when $\alpha$ is fixed at 20. Smaller $k$ values cause $\alpha_t$ to decay more slowly over time, weakening constraints from the original image and facilitating more complete lesion removal. Conversely, larger $k$ values strengthen constraints from the original image, which may hinder full lesion elimination. Tab.~\ref{tab:hyperparam_acg} quantitatively corroborates this trend: moderate $k$ values achieve a balance between preserving anatomical consistency and effectively removing lesions, providing an optimal trade-off between quantitative performance and qualitative outcomes. Based on these observations, we set $k=0.5$ as the default value.

\vspace{0.1cm}
\noindent \textbf{Lesion Consistency Guidance (LCG).}
For LCG, we perform an ablation study on the healthy-to-pathology editing task by varying the hyperparameter $\eta$. 
Fig.~\ref{fig:hyperpara_lcg} illustrates that $\eta$ controls the strength of the lesion guidance: smaller values of $\eta$ weaken the guidance, resulting in less pronounced lesion regions in the edited pathological images. 
Given our design, $\eta$ is intended to be less than or equal to 1; values greater than 1 (e.g., $\eta=2$) can lead to over-erosion of the lesion regions. 
Based on these observations, we adopt $\eta=1$ as the default setting in all experiments.

%%%%%%%%%%%%%%%%%%%%%%%%%%%%%%%%%%%%%%
%%%%%%%%%%%%%%%%%%%%%%%%%%%%%%%%%%%%%%

\section{Further Discussion and Broader Impact}
\label{suppl:exp_real}

\input{fig_tex/fig_comparison_atlas}
\input{fig_tex/fig_comparison_atlas_artifact}

In Sec.~\ref{subsec:editing_exp} and Suppl.~\ref{suppl:editing_exp}, we evaluate \texttt{USB}'s bidirectional pathology-to-healthy editing performance, using synthetic, paired pathology-healthy images --- which allows for both qualitative and quantitative evaluation.
In this section, we conduct further experiments and discussion on demonstrating \texttt{USB}’s editing robustness in challenging clinical scenarios, where \textit{no ground-truth exists}.

In particular, we perform pathology-to-healthy editing on: 
\textit{(i)}~real stroke patient scans to reconstruct their healthy counterparts;  
\textit{(ii)}~real-world scans corrupted during acquisition.

\vspace{0.1cm}
\noindent \textbf{Editing as Healthy Reconstruction for Patient Scans.} 
Fig.~\ref{fig:comparison_atlas} illustrates the results of different methods applied to real stroke pathological brain images from the ATLAS dataset~\cite{Liew2017ATLAS}. 
As shown, for relatively small lesions (first row), competing methods can partially reconstruct healthy brain structures, yet the results are noticeably inferior to those produced by \texttt{USB}. 
For larger lesions (second row), all other methods fail to recover the healthy regions, whereas \texttt{USB} successfully restores anatomically plausible healthy brain structures. 
These observations demonstrate \texttt{USB}'s superior ability to handle a wide range of lesion sizes and its robustness in generating anatomically plausible healthy reconstructions from real pathological scans, even though the model was trained solely on pseudo-synthetic pathological data.

\vspace{0.1cm}
\noindent \textbf{Editing as Image Inpainting for Corrupted Scans.}  
%%%%%%%%%%%%% artifact %%%%%%%%%%%%%%%%
In clinical practice, various factors (e.g., patient motion, metallic implants, bias fields) can introduce artifacts and corruptions in the acquired MRI scans. As illustrated in Fig.~\ref{fig:comparison_atlas_artifact}, such artifacts may occlude large brain regions, posing additional challenges for downstream image analysis. 
Experimental results indicate that other methods fail to handle such occluded regions, leaving the artifacts uncorrected in the edited images. In contrast, \texttt{USB} demonstrates strong inpainting capabilities, effectively reconstructing the lesioned areas while simultaneously restoring regions affected by imaging artifacts. This highlights \texttt{USB}'s robustness in editing real pathological scans even under challenging conditions, further validating its ability to generalize from pseudo-synthetic pathological training data to real clinical images.

%%%%%%%%%%%%%%%%%%%%%%%%%%%%%%%%%%%%%%
%%%%%%%%%%%%%%%%%%%%%%%%%%%%%%%%%%%%%%

\section{Limitations}
\label{suppl:limitation}

Due to the scarcity of paired pathology-healthy scans, \texttt{USB} lacks ground-truth references for evaluating the clinical correctness of synthesized healthy or pathological structures. While we resort to quantitative metrics including FID, KID, and MMD, to assess image fidelity, they cannot fully capture clinical plausibility or diagnostic interpretability. As an important next step, we plan to conduct systematic, structured reader studies in collaboration with radiologists and neurologists to evaluate the synthesis quality of \texttt{USB}'s generations, especially in terms of their anatomical realism and clinical utility, thus enabling a more rigorous and clinically meaningful assessment of the framework.

%%%%%%%%%%%%%%%%%%%%%%%%%%%%%%%%%%%%%%
%%%%%%%%%%%%%%%%%%%%%%%%%%%%%%%%%%%%%%

\section{Datasets and Metrics}
\label{suppl:dataset}

\subsection{Datasets and Preprocessing}

\begin{itemize}

\item \texttt{HCP}~\cite{Essen2012TheHC}: we use T1-weighted (897 cases) MRI scans of young subjects from the Human Connectome Project, acquired at 0.7 mm resolution.

\item \texttt{ADHD200}~\cite{Brown2012ADHD200GC}: we use T1-weighted (961 cases) MRI scans from the ADHD200 Sample, which is a grassroots initiative dedicated to the understanding of the neural basis of Attention Deficit Hyperactivity Disorder (ADHD).

\item \texttt{OASIS3}~\cite{LaMontagne2018OASIS3LN}: we use MRI (885 cases) scans from OASIS3, which is a longitudinal neuroimaging, clinical, and cognitive dataset for normal aging and AD. For our experiments.

\item \texttt{ATLAS}~\cite{Liew2017ATLAS}: we use provided gold-standard stroke lesion segmentations (655 cases), from Anatomical Tracings of Lesions After Stroke (ATLAS), which is a study of subacute/chronic stroke.

\item \texttt{ISLES}~\cite{Hernandez2022ISLES} we use the provided gold-standard stroke lesion segmentation (152 cases) from ISLES 2022, which is a MICCAI challenge in 2022 for acute/subacute stroke lesion detection and segmentation.

\item \texttt{ADNI}~\cite{Jack2008TheAD}: we use T1-weighted (2045 cases) MRI scans from the Alzheimer’s Disease Neuroimaging Initiative (ADNI). All scans are acquired at 1 $mm$ isotropic resolution from a wide array of scanners and protocols. The dataset contains aging subjects, some diagnosed with mild cognitive impairment (MCI) or Alzheimer's Disease (AD). Many subjects present strong atrophy patterns and white matter lesions.

\end{itemize}

Among the above six datasets, \texttt{HCP}~\cite{Essen2012TheHC}, \texttt{ADHD200}~\cite{Brown2012ADHD200GC}, \texttt{OASIS3}~\cite{LaMontagne2018OASIS3LN} contain subjects with healthy anatomy. \texttt{ATLAS}~\cite{Liew2017ATLAS}, \texttt{ISLES}~\cite{Hernandez2022ISLES}. \texttt{ATLAS} and \texttt{ISLES} include stroke patients, with gold-standard manual segmentations of stroke lesions provided in both datasets. \texttt{ADNI}~\cite{Jack2008TheAD} consists of Alzheimer's disease patients, most of whom exhibit white matter hyperintensities (WMH). We obtain the WMH labels using LST-AI~\cite{wiltgen2024lst}, which provides silver-standard segmentations.

\subsection{Metrics}

We resort to various metrics for evaluating individual tasks across multiple aspects: 

\begin{itemize}
\item \texttt{FID}: the Fr\'{e}chet Inception Distance, which measures the distributional distance between real and generated images by comparing their feature statistics (mean and covariance) extracted from a pretrained Inception network. Following the 3D-FID protocol of~\cite{sun2022hierarchical}, we use a pretrained 3D ResNet for feature extraction, ensuring that the metric reflects volumetric structural fidelity in brain MRI data.

\item \texttt{KID}: the Kernel Inception Distance, it computes the squared Maximum Mean Discrepancy (MMD) between Inception features of real and generated images using a polynomial kernel.

\item \texttt{MMD}: the Maximum Mean Discrepancy, which measures the distance between two distributions in a reproducing kernel Hilbert space (RKHS).

\item \texttt{L1}: the average $L1$ distance, which measures the voxel-wise reconstruction accuracy and is widely used to assess anatomy recovery quality.

\item \texttt{PSNR}: the Peak Signal-to-Noise Ratio, which reflects the fidelity of reconstructed images relative to the ground truth.

\item \texttt{SSIM}: the Structural Similarity Index, which evaluates perceived structural consistency between generated and real images, focusing on luminance, contrast, and structural alignment, making it well-suited for assessing anatomical reconstruction quality.
\end{itemize}

\section{Preliminary: Standard Diffusion}
\label{suppl:preliminary}

For completeness, we briefly overview the standard diffusion framework~\cite{rombach2022ldm}, which consists of a \emph{forward} (noising) process and a \emph{reverse} (denoising) process.

\vspace{0.1cm}
\noindent \textbf{Forward Process.}
Given a data sample $x_0 \sim q(x_0)$, the forward diffusion process gradually adds Gaussian noise over $T$ steps, defined as a Markov chain:
%\vspace{-0.2cm}
\begin{equation}
q(x_t \mid x_{t-1}) = \mathcal{N}\big(x_t; \sqrt{1-\beta_t} \, x_{t-1}, \, \beta_t \mathbf{I}\big),
%\vspace{-0.2cm}
\end{equation}
where $\beta_t \in (0,1)$ is a pre-defined noise schedule, and we define $\alpha_t = 1 - \beta_t$ as the per-step signal preservation factor.  
The cumulative product $\bar{\alpha}_t = \prod_{s=1}^{t} \alpha_s$ represents the total amount of signal retained after $t$ forward steps.

The closed-form marginal distribution of $x_t$ given $x_0$ is:
%\vspace{-0.2cm}
\begin{equation}
q(x_t \mid x_0) = \mathcal{N}\big(x_t; \sqrt{\bar{\alpha}_t} \, x_0, \, (1-\bar{\alpha}_t) \mathbf{I}\big),
%\vspace{-0.2cm}
\end{equation}
where $\bar{\alpha}_t$ controls the decay of the original data content through the noising process.
After $T$ steps, the data is transformed approximately into isotropic Gaussian noise: $x_T \sim \mathcal{N}(0, \mathbf{I})$.

\vspace{0.1cm}
\noindent \textbf{Reverse Process.}
The reverse (generative) process aims to invert the forward diffusion:
%\vspace{-0.2cm}
\begin{equation}
p_\theta(x_{t-1} \mid x_t) = \mathcal{N}\big(x_{t-1}; \mu_\theta(x_t, t), \, \Sigma_\theta(x_t, t)\big),
%\vspace{-0.2cm}
\end{equation}
where $\mu_\theta$ and $\Sigma_\theta$ are predicted by a neural network with parameters $\theta$. 
The joint probability of the reverse process can be written as:
%\vspace{-0.3cm}
\begin{equation}
p_\theta(x_{0:T}) = p(x_T) \prod_{t=1}^{T} p_\theta(x_{t-1} \mid x_t),
%\vspace{-0.3cm}
\end{equation}
with the prior $p(x_T)=\mathcal{N}(0,\mathbf{I})$.

\vspace{0.1cm}
\noindent \textbf{Training Objective}. 
Training minimizes the variational upper bound (ELBO) between forward and reverse processes, which is simplified to a noise-prediction loss in practice:
%\vspace{-0.2cm}
\begin{equation}
\mathcal{L}_{\text{simple}} = \mathbb{E}_{x_0, \epsilon, t} \Big[ \big\| \epsilon - \epsilon_\theta(\sqrt{\bar{\alpha}_t} x_0 + \sqrt{1-\bar{\alpha}_t} \epsilon, t) \big\|^2 \Big],
%\vspace{-0.2cm}
\end{equation}
where $\epsilon \sim \mathcal{N}(0, \mathbf{I})$ and $\epsilon_\theta$ denotes the predicted noise.

\end{document}